\documentclass[10pt,twocolumn,letterpaper]{article}
\usepackage[pagenumbers]{cvpr}
\usepackage[T1]{fontenc}
\usepackage{iftex}
\ifPDFTeX
  \usepackage[utf8]{inputenc}
\fi
\usepackage{bm}
\usepackage{tabularx,array}
\usepackage{enumitem}
\usepackage{float}
\usepackage{needspace}
\usepackage{microtype}
\newcolumntype{Y}{>{\raggedright\arraybackslash}X}

\setlist{itemsep=3pt,topsep=5pt,parsep=0pt}
\renewcommand{\arraystretch}{1.16}
\definecolor{cvprblue}{rgb}{0.21,0.49,0.74}
\usepackage[pagebackref,breaklinks,colorlinks,allcolors=cvprblue]{hyperref}
\usepackage{xurl}
\def\confName{CVPR}
\def\confYear{2027}
\hypersetup{pdfauthor={Yutian Zhang, Siyuan Ma, Runhan Huang, Hanyu Zhu, Yudong Gao, Boshi Zhang, Fenghua Zhang, Yuhang Ming, Haozhen Chi, Dibo Hou, Xiaojin Huang, Qiaojun Yu},pdftitle={WB-RSI: Recursive Self-Improvement for Humanoid Whole-Body Manipulation through World-Model Planning and Active Perception}}

\title{WB-RSI: Recursive Self-Improvement for Humanoid Whole-Body Manipulation through World-Model Planning and Active Perception}
\author{%
Yutian Zhang$^{1,*}$, Siyuan Ma$^{2,*}$, Runhan Huang$^{2,*}$, Hanyu Zhu$^{1}$, Yudong Gao$^{3}$, Boshi Zhang$^{2}$,\\
Fenghua Zhang$^{4}$, Yuhang Ming$^{5}$, Haozhen Chi$^{1}$, Dibo Hou$^{1}$, Xiaojin Huang$^{2}$, Qiaojun Yu$^{6}$\\[4pt]
{\small $^{1}$Zhejiang University \quad $^{2}$Tsinghua University}\\
{\small $^{3}$The Hong Kong University of Science and Technology \quad $^{4}$University of Southern California}\\
{\small $^{5}$Shanghai Jiao Tong University \quad $^{6}$Shanghai Artificial Intelligence Laboratory}\\[3pt]
{\footnotesize $^{*}$Equal contribution.}\\[4pt]
{\small Project Page: \url{https://ytydt-reuz.github.io/WB-RSI-Website/}}
}

\begin{document}
\maketitle
\begin{abstract}
Humanoid robots deployed for whole-body manipulation should improve from their own experience rather than wait for offline retraining. Achieving such recursive self-improvement (RSI) on humanoids requires accumulating and reusing verified experience across deployments, anticipating the physical consequences of whole-body actions before committing to them, and actively acquiring missing observations before contact. We present \textbf{WB-RSI}, an agentic self-evolving framework for humanoid loco-manipulation built around an evolving agentic harness that unifies experience memory, reusable skills, world-model planning, and active perception. During execution, an agentic orchestrator schedules skill reuse, predictive planning, active observation, and failure recovery over a shared whole-body command interface; between deployment rounds, verified successes, recoveries, and failure guards consolidate into memory and skills without updating model weights. World-model planning forecasts visual outcomes to select effective actions locally and offload routine foundation-model queries, while active perception coordinates body motions to bring initially unseen targets into view and verify task geometry before acting. In simulation, WB-RSI continually improves across deployment rounds, substantially boosting task completion while cutting foundation-model calls and token costs. On a physical Unitree G1 humanoid, WB-RSI achieves 90\% mean success across five whole-body loco-manipulation tasks, outperforming leading VLA, WAM, and agentic baselines.

\end{abstract}
\begin{figure*}[t]
\centering
\includegraphics[width=0.90\textwidth]{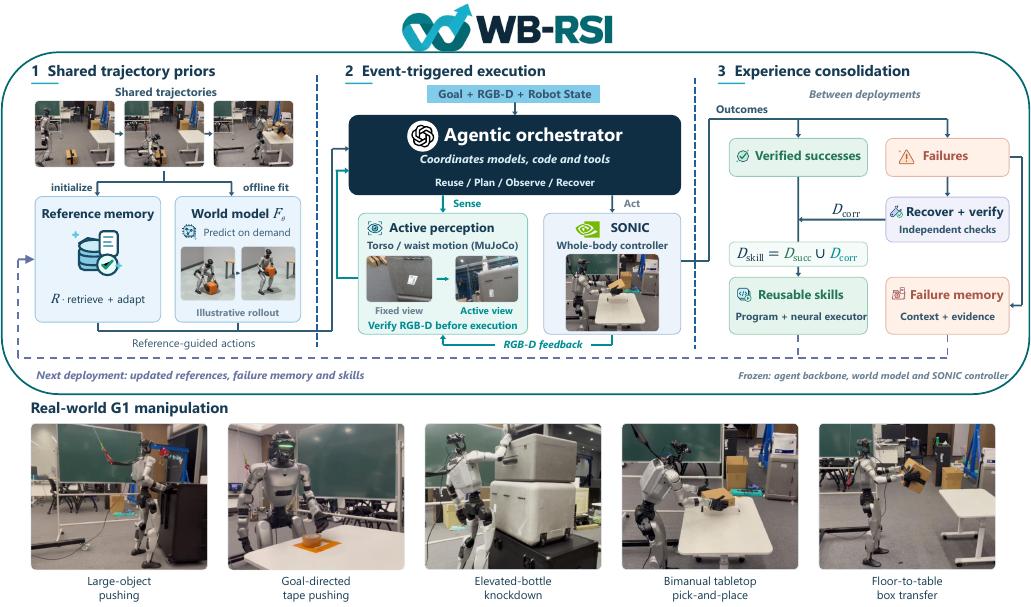}
\caption{\textbf{Overview of WB-RSI.} Offline trajectories initialize the experience memory and fit the action-conditioned world model. During deployment, an agentic orchestrator coordinates experience reuse, world-model planning, active perception, and whole-body execution. Between deployment rounds, verified successes, recoveries, and failure guards consolidate into memory and reusable skills, while the foundation-model agent, world model, and whole-body controller remain frozen. Bottom: five real-world Unitree G1 manipulation tasks.}
\label{fig:framework}
\end{figure*}

\section{Introduction}\label{introduction}

Humanoid robots are progressing from robust locomotion \cite{zhuang2024humanoid,zhu2026hiking} to whole-body loco-manipulation \cite{weng2025hdmi,zhao2025resmimic,dong2026hero,li2026flore}, where the legs, torso, arms, and hands must coordinate to interact with objects across changing scenes. Whole-body controllers \cite{he2024omnih2o,fu2024humanplus,he2024hover,luo2025sonic,ye2026skillx} paired with generalist policies, ranging from whole-body vision-language-action (VLA) models \cite{jiang2025wholebodyvla,wei2026psi0,hu2026openhlm,bjorck2025gr00t, FWBC} to world action models (WAMs) \cite{qin2026wbwam,zhang2026wholebodywam,zheng2026motionwam,li2026omega0,DECOWAM}, now provide strong motor foundations, yet their competence remains static once deployed unless new demonstrations and retraining are supplied. Foundation-model agents offer a complementary paradigm: they connect high-level reasoning to robot control \cite{ahn2022saycan,liang2023code,huang2023voxposer,yuan2025being0,ren2026cybowaiter,choi2026hugo} and can self-improve by accumulating episodic memory, distilling failure feedback, and expanding skill libraries without weight updates \cite{shinn2023reflexion,zhao2024expel,wang2023voyager,huang2026rax,lu2026aspire,xiao2026enpire}. Realizing recursive self-improvement (RSI) in humanoid loco-manipulation, however, demands three tightly coupled capabilities. First, the agentic harness must continually evolve and remain disciplined during execution: verified outcomes should refine memory and skills across rounds while an orchestrator invokes the foundation model only at genuine decision points. Second, because whole-body actions are high-dimensional, costly to reverse, and contact-sensitive, the robot must anticipate action consequences and evaluate candidate plans before committing to execution \cite{hafner2023dreamer,du2023unipi}. Third, because manipulation targets frequently start outside the egocentric field of view, the robot must actively adjust its body posture to acquire and verify task-critical observations before acting \cite{xiong2025vision}.

We present \textbf{WB-RSI}, a self-evolving framework for humanoid loco-manipulation built around these capabilities (Figure~\ref{fig:framework}). Its core is an evolving harness operating over a unified whole-body command interface, where an orchestrator coordinates skill reuse, planning, observation, and recovery, consulting the foundation model only when novel decisions arise or execution deviates from expectation. To ground these decisions in the physical scene, active perception adjusts the robot's torso and body posture to bring initially out-of-view targets into sight and confirm their spatial state before contact. Building on these observations, world-model planning forecasts the visual outcomes of candidate whole-body actions so the robot can evaluate progress and physical risk before acting, selecting effective maneuvers locally and avoiding costly trial-and-error. Between deployment rounds, validated successes, recoveries, and failure records update an experience memory and consolidate into reusable skills, allowing the robot to avoid past mistakes and turn behaviors that once required expensive planning or foundation-model queries into fast, reliable reuse in subsequent rounds without updating any model weights.

We evaluate WB-RSI across simulation benchmarks and on a physical Unitree G1 \cite{unitreeg1}. In simulation, evolving from deployment experience substantially improves multi-round task completion and execution speed, reaching 100\% held-out success after ten closed-loop rounds while cutting foundation-model calls by 85.7\% compared with a frozen harness. Both world-model planning and active perception contribute clear gains, cutting multi-step LLM token cost by over half, reliably verifying initially out-of-view targets, and improving out-of-distribution generalization over prior agentic methods across both closed- and open-weight backbones. On the physical Unitree G1, WB-RSI achieves 90\% mean success across five whole-body loco-manipulation tasks, outperforming leading agentic, WAM, and VLA baselines by 9 to 19 percentage points. Videos and additional results are available on our project page: \url{https://ytydt-reuz.github.io/WB-RSI-Website/}.

Our main contributions are threefold:
\begin{itemize}
\item \textbf{Self-Evolving Whole-Body Harness:} We introduce \textbf{WB-RSI}, a recursive self-improvement framework for humanoid loco-manipulation that continually turns deployment experience, including both successes and failures, into reusable memory and skills across rounds without updating any model or controller weights.
\item \textbf{Prediction- and Perception-Gated Orchestration:} We combine world-model planning and body-driven active perception over a unified whole-body command interface, resolving out-of-view targets prior to contact and offloading routine action selection from costly foundation-model queries.
\item \textbf{Simulation and Physical Humanoid Validation:} We demonstrate substantial gains in multi-round self-improvement, query efficiency, and out-of-distribution generalization in simulation, and achieve 90\% mean success across five physical loco-manipulation tasks on a Unitree G1 against nine VLA, WAM, and agentic baselines.
\end{itemize}

\section{Related Work}\label{related-work}

\subsection{Humanoid Loco-Manipulation}\label{related-humanoid}

Learned whole-body controllers give humanoids a general motion interface. OmniH2O~\cite{he2024omnih2o} and HumanPlus~\cite{fu2024humanplus} learn from human demonstrations, HOVER~\cite{he2024hover} and SONIC~\cite{luo2025sonic} track diverse whole-body commands at scale, and ResMimic~\cite{zhao2025resmimic}, HERO~\cite{dong2026hero}, and force-reactive control~\cite{li2026flore} extend tracking to object interaction, visual grasping, and compliant contact. On top of such interfaces, manipulation behavior is commonly learned from demonstrations, either with generalist vision-language-action policies \cite{brohan2023rt2,kim2024openvla,black2024pi0,physicalintelligence2025pi05} and their whole-body humanoid counterparts \cite{bjorck2025gr00t,jiang2025wholebodyvla,wei2026psi0,hu2026openhlm}, with action-sequence models \cite{chi2023diffusion,zhao2023act}, or with world action models that jointly predict video and actions \cite{ye2026dreamzero,yuan2026fastwam}, which recent work extends to humanoid loco-manipulation \cite{qin2026wbwam,zhang2026wholebodywam,zheng2026motionwam,li2026omega0}. Demonstrations can also be reused directly through retrieval, in-context conditioning, or skill discovery \cite{pari2022vinn,memmel2024strap,fu2024icrt,zhu2021buds}. While effective within their training coverage, improving these policies after deployment typically requires collecting additional supervision or updating weights via interactive imitation or learned search \cite{ross2011dagger,jain2025sailor,yang2026eqc}. WB-RSI builds on a frozen whole-body controller and improves task-level competence directly from deployment experience, without retraining the controller or foundation-model backbone.

\subsection{Agentic Robotics and Recursive Self-Improvement}\label{related-agentic-rsi}

Foundation-model agents connect language reasoning to robot actions by selecting skills \cite{ahn2022saycan}, incorporating execution feedback \cite{huang2022inner,yao2022react}, writing control programs and spatial objectives \cite{liang2023code,huang2023voxposer}, and reasoning over embodied observations \cite{driess2023palme}; HITCH~\cite{lu2026hitch} brings this paradigm to humanoids through a whole-body harness. Language agents further self-improve without weight updates: Voyager~\cite{wang2023voyager} grows an executable skill library, while Reflexion~\cite{shinn2023reflexion} and ExpeL~\cite{zhao2024expel} distill execution feedback into reusable guidance. Recent robot agents carry this loop into physical interaction: RAX~\cite{huang2026rax} scales online exploration into memory and verified skills, ASPIRE~\cite{lu2026aspire} iteratively synthesizes and refines robot control programs, and ENPIRE~\cite{xiao2026enpire} improves real-world manipulation policies through agentic feedback. Beyond accumulating past trajectories, a self-improving humanoid agent must anticipate the physical consequences of whole-body commands and actively acquire missing visual evidence. World models address the former through latent dynamics planning \cite{hafner2019planet,hafner2023dreamer,hansen2024tdmpc2}, video simulation \cite{du2023unipi,yang2023unisim}, and action-conditioned outcome evaluation for robot agents \cite{zhang2026wap,gao2026imwm,wang2026embodiedskills}, with EvoWorld~\cite{zhang2026evoworld} further leveraging world-model rollouts to orchestrate and continuously evolve modular embodied skills; active perception addresses the latter through purposeful viewpoint control \cite{xiong2025vision}. WB-RSI unifies both within an evolving whole-body harness, where active perception gates execution on verified evidence and predictive planning filters actions before outcomes are consolidated across rounds.

\section{Method}\label{wbc-rsi-overview}

WB-RSI equips a humanoid with a self-evolving harness that improves across deployment rounds (Figure~\ref{fig:framework}). During execution, an orchestrator coordinates experience memory \(\mathcal R\), a reusable skill library \(\mathcal K\), world-model planning, and active perception, calling a foundation-model agent only when local memory and predictions cannot decide the next action. Memory and skills form the evolving part of the harness in round \(r\), written \(\mathcal H^r=(\mathcal R^r,\mathcal K^r)\). At step \(t\), given task goal \(g\), camera observation \(o_t\), and measured robot state \(x_t\), every module outputs actions in one shared format: a short-horizon command block \(U_t=(u_t,\ldots,u_{t+H-1})\) executed by a frozen whole-body controller. After round \(r\), recorded experience \(\mathcal D^r\) updates the harness:
\begin{equation}
\mathcal H^{r+1}=\operatorname{Update}\big(\mathcal H^r,\mathcal D^r\big).
\label{eq:harness-update}
\end{equation}
Section~\ref{sec:method-harness} describes the command interface, orchestrator, and cross-round updates of \(\mathcal H^r\); Sections~\ref{sec:method-wm} and~\ref{sec:method-ap} detail world-model planning and active perception.

\subsection{Agentic Orchestrator and Self-Evolving Harness}\label{sec:method-harness}

\begin{figure}[t]
\centering
\includegraphics[width=0.80\columnwidth]{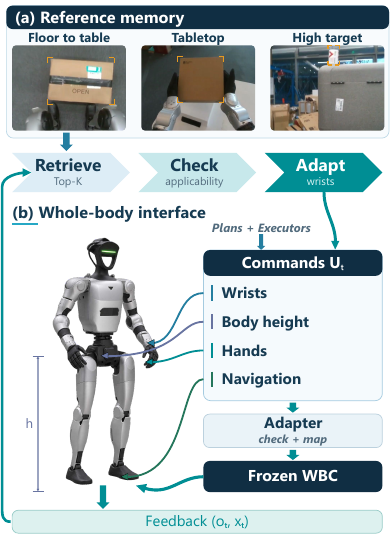}
\caption{\textbf{Reference reuse and whole-body interface.} (a) Retrieved references are adapted via object-relative wrist commands. (b) All modules output shared command blocks $U_t$ mapped to the frozen whole-body controller (WBC), with sensor feedback $(o_t, x_t)$ closing the loop.}
\label{fig:whole_body_interface}
\end{figure}

\paragraph{Whole-body command interface.}
All modules control the robot through one shared whole-body command interface (Figure~\ref{fig:whole_body_interface}(b)). Each command block \(U_t\) combines wrist poses, hand states, body height, and base motion. Before execution, a command adapter checks joint and workspace limits, converts coordinates, and interpolates accepted blocks for the low-level whole-body controller~\cite{luo2025sonic}. Because retrieved references, world-model plans, skills, and observation motions all use the same command format \(U_t\), the orchestrator can switch smoothly between them and log every action uniformly. Timestamp checks discard outdated commands, and issued commands are logged separately from measured robot states \(x_t\), so the world model predicts from commands while sensor readings confirm what actually happened.

\paragraph{Orchestration.}
During execution, the orchestrator switches among six modes: \texttt{ASSESS} reads current sensors and searches \(\mathcal R\) and \(\mathcal K\); \texttt{REUSE} runs a matching skill or adapted reference; \texttt{PLAN} invokes world-model planning; \texttt{OBSERVE} runs active perception; \texttt{RECOVER} fixes a failed step; and \texttt{STOP} ends the episode. Transitions follow a fixed priority order of \texttt{STOP}, \texttt{RECOVER}, \texttt{OBSERVE}, \texttt{REUSE}, and \texttt{PLAN}, so the robot resolves missing views before touching objects and reuses known skills before searching for new actions. Between agent calls, sensor feedback drives these transitions automatically. If execution drifts from its target or stops making progress, the orchestrator triggers \texttt{PLAN}, \texttt{RECOVER}, or a foundation-model query within fixed step and time limits.

\paragraph{Experience memory.}
The experience memory \(\mathcal R\) stores past execution segments split at contact transitions, task phases, and completion events, recording each segment's context, command block \(U_t\), and local outcome. Successful task segments \(\mathcal D_{\mathrm{succ}}\) and successful recovery segments \(\mathcal D_{\mathrm{corr}}\) serve as reusable references. At test time, retrieval matches candidates by goal, visual scene, robot posture, and task phase, then adapts their wrist trajectories to the current object pose through object-relative transformations (Figure~\ref{fig:whole_body_interface}(a)). Failed attempts are stored in \(\mathcal B_{\mathrm{fail}}\subseteq\mathcal R\) together with their context, attempted commands, and observed failure events, preventing the robot from repeating actions that previously failed under similar conditions and guiding recovery. Crucially, outcomes are labelled per segment rather than per episode: a good approach remains reusable even if a later grasp fails, while the failed grasp keeps its negative label separately from any subsequent recovery.

\paragraph{Skill library.}
Procedures that succeed repeatedly are turned into reusable skills in \(\mathcal K\). Each skill combines a short control program, which defines when the skill can start, what it must observe, and how to tell if it succeeded or failed, with a low-level executor, implemented either as parameterized primitive routines in the style of Code as Policies~\cite{liang2023code} or as a short-horizon neural policy \(\pi_\phi\)~\cite{chi2023diffusion}. Programs write actions in object-relative coordinates over a fixed tool interface, and recovery branches are added only after they succeed in real execution. Both task successes and successful recoveries provide data for consolidating skills:
\begin{equation}
\mathcal D_{\mathrm{skill}}=\mathcal D_{\mathrm{succ}}\cup\mathcal D_{\mathrm{corr}}.
\label{eq:skill-data}
\end{equation}

\paragraph{Cross-deployment evolution.}
After each deployment round, Eq.~\eqref{eq:harness-update} adds new successful and failed segments to \(\mathcal R\), updates skill programs and executors using \(\mathcal D_{\mathrm{skill}}\), and tests each updated harness on a regression set before deploying it. Throughout this process, the foundation-model agent, world model, and whole-body controller stay frozen. As \(\mathcal R^r\) and \(\mathcal K^r\) grow across rounds, cases that first required planning or foundation-model queries gradually become fast, direct skill reuse, lowering reasoning cost while steadily improving reliability.

\subsection{World-Model Planning}\label{sec:method-wm}

World-model planning lets the robot foresee how candidate actions will look and play out before touching an object. Following action-conditioned video prediction~\cite{zhang2026wap}, a frozen visual world model \(p_\theta\) trained offline on past interactions predicts future camera frames given the image history \(o_{\le t}\), past commands \(u_{<t}\), and a proposed command block \(U_t\):
\begin{equation}
\hat o_{t+1:t+H}\sim p_\theta\big(\cdot\mid o_{\le t},u_{<t},U_t\big).
\label{eq:world-model}
\end{equation}
Because \(p_\theta\) takes the same command format used on the robot, its predictions reflect how the whole-body controller actually responds.

In \texttt{PLAN}, the orchestrator builds a small set of candidate actions \(\mathcal U_t\) from task primitives and local variations of retrieved references, dropping any candidate that violates workspace limits or matches a known failure in \(\mathcal B_{\mathrm{fail}}\). For each remaining candidate \(U\in\mathcal U_t\), the world model predicts a rollout \(\hat o(U)\), and a task scorer estimates its progress \(S_{\mathrm{task}}(\hat o(U),g)\), such as how high an object is lifted during grasping or how much closer it moves to the goal during transport. Candidates whose predicted frames keep the target clearly tracked and show a clear progress margin form the supported set \(\mathcal U_t^{\mathrm{supp}}\). Whenever \(\mathcal U_t^{\mathrm{supp}}\) is nonempty, the orchestrator picks the best candidate locally without calling the agent:
\begin{equation}
U^\star=\arg\max_{U\in\mathcal U_t^{\mathrm{supp}}}
S_{\mathrm{task}}\big(\hat o(U),g\big).
\label{eq:wm-selection}
\end{equation}
If no candidate passes these checks, the orchestrator asks the foundation-model agent to choose using current images and episode history. After execution, real sensor readings check whether the action succeeded.

\subsection{Active Perception}\label{sec:method-ap}

In whole-body manipulation, targets such as floor-level boxes or bottles on high shelves often start outside the robot's camera view. \texttt{OBSERVE} solves this before contact by using whole-body viewpoint motions, including waist pitch and yaw, body-height changes, lateral scans, and paused re-measurements, through the same command interface. Floor targets trigger downward scans, while elevated targets combine upward tilt with side-to-side sweeps.

Active perception proceeds in two stages. First, the robot searches across predefined viewpoints and applies small RGB centering adjustments to bring the target cleanly into view. Second, it checks task-specific physical conditions from live RGB-D measurements, such as target position before approach, two-handed contact before lifting, or table clearance before placement. Once confirmed, control returns to \texttt{ASSESS} and manipulation proceeds; otherwise, execution pauses until valid observations are obtained or the search budget runs out. Successful viewpoints are saved in \(\mathcal R\) with their task context, allowing later rounds to locate targets with fewer scans.

\section{Experiments}\label{experiments}

\subsection{Experimental Setup}\label{platform-and-implementation}
We evaluate WB-RSI in simulation and on a physical Unitree G1~\cite{unitreeg1}, testing cross-deployment self-improvement, world-model planning, active perception, out-of-distribution generalization, and end-to-end physical performance. Simulation studies pair fixed-base G1 RGB-D active-perception environments with MuJoCo contact-proxy manipulation tasks (box transfer, bottle knockdown, and large-object pushing) built on RAX~\cite{huang2026rax} with GPT-6.1 Sol as the default agent. Physical evaluation covers five whole-body loco-manipulation tasks on the G1: large-object pushing, goal-directed tape pushing, elevated-bottle knockdown, bimanual tabletop pick-and-place, and floor-to-table box transfer, denoted T1--T5. Within each study, all methods share matched scene layouts; complete protocols appear in the supplement. Unless noted otherwise, reported costs average all attempts including failures, tokens count input-plus-output usage in thousands, online wall time covers sensing, execution, world-model rollouts, and agent inference, and $T_B$ assigns completion time to successes and budget $B$\,s to failures. Boldface highlights the highest success and the lowest cost among top-performing methods.

\subsection{Self-Evolving Harness}\label{closed-loop-validation}\label{rsi-multistep-validation}

We first examine how the harness improves task competence and efficiency from its own deployments.

\begin{table}[t]
\centering\small
\setlength{\tabcolsep}{1.2pt}
\caption{Closed-loop box transfer after ten deployment rounds (30 held-out tests per configuration; Frozen memory token mean over 28 complete-usage tests, 30 otherwise).}
\label{tab:closed_loop_ablation}
\begin{tabular*}{\columnwidth}{@{}l@{\extracolsep{\fill}}rrrrr@{}}
\toprule
Configuration & SR (\%) $\uparrow$ & Calls $\downarrow$ & Scans $\downarrow$ & Tok. (k) $\downarrow$ & $T_{180}$ (s) $\downarrow$ \\
\midrule
w/o AP & 30.0 & 0.00 & 0.00 & 0.00 & 128.50 \\
w/o WM planning & 90.0 & 0.20 & 0.70 & 9.81 & 23.17 \\
Frozen memory & 93.3 & 0.70 & 0.70 & 12.08 & 51.90 \\
WB-RSI (ours) & \textbf{100.0} & \textbf{0.10} & \textbf{0.70} & \textbf{1.76} & \textbf{12.33} \\
\bottomrule
\end{tabular*}
\end{table}

\paragraph{Closed-loop self-improvement.}
Across three independent evolution streams, WB-RSI runs ten deployment rounds of box transfer on randomized layouts, varying object pose, geometry, dynamics, goal location, and initial camera and gripper states, and consolidates verified outcomes into memory after each round. Checkpoints are evaluated on 30 locked held-out layouts that never update memory, comparing WB-RSI against frozen memory, disabled active perception, and an ablation that replaces world-model planning with local RGB-geometry selection under identical routing.

As shown in Table~\ref{tab:closed_loop_ablation}, ten rounds of evolution bring WB-RSI to 100\% held-out success while averaging just 0.10 agent calls per task, surpassing frozen memory while cutting agent queries by 85.7\% and slashing $T_{180}$ by over three-quarters. Removing active perception drops completion to 30.0\%, as the robot can only solve initially visible targets, whereas replacing world-model planning during evolution lowers final success to 90.0\% and nearly doubles $T_{180}$. Notably, by round 10, high-value grasp and viewpoint choices discovered through predictive rollouts during deployment have been distilled into \(\mathcal R\), enabling WB-RSI to solve all 30 held-out tasks with zero online world-model batches and minimal agent intervention.

\begin{table}[t]
\centering\footnotesize
\setlength{\tabcolsep}{1.3pt}
\caption{Experience consolidation and skill reuse: (a) ten-round memory ablation (60 matched tasks) and (b) checked-program reuse (24 new paired layouts). S/F denote positive/negative outcome records.}
\label{tab:rsi_multistep_results}
\begin{tabular*}{\columnwidth}{@{}l@{\extracolsep{\fill}}rrrrr@{}}
\toprule
Configuration & Success $\uparrow$ & Calls $\downarrow$ & Scans $\downarrow$ & Tok. (k) $\downarrow$ & Online (s) $\downarrow$ \\
\midrule
\multicolumn{6}{@{}l}{(a) Ten-round mechanism ablation (60 matched tasks)} \\
No episodic memory & 20/60 & 4.17 & 3.00 & 97.13 & 70.20 \\
Fixed initial memory & 22/60 & 4.33 & 2.17 & 102.33 & 68.53 \\
S updates & 38/60 & 3.50 & \textbf{2.00} & 84.66 & 43.78 \\
S+F evidence & 48/60 & 3.50 & \textbf{2.00} & 84.95 & 34.27 \\
WB-RSI (ours) & \textbf{51/60} & \textbf{3.17} & 2.17 & \textbf{76.16} & \textbf{27.10} \\
\midrule
\multicolumn{6}{@{}l}{(b) Checked-program reuse (24 new paired layouts)} \\
S+F evidence & \textbf{24/24} & 3.54 & \textbf{2.04} & 87.12 & 105.17 \\
WB-RSI (ours) & \textbf{24/24} & \textbf{3.08} & 2.08 & \textbf{74.97} & \textbf{88.88} \\
\bottomrule
\end{tabular*}
\end{table}

\paragraph{What to remember.}
To disentangle how memory content drives self-improvement, we evaluate five variants across ten rounds of box transfer and bottle knockdown on 60 matched tasks with world-model planning disabled. As reported in Table~\ref{tab:rsi_multistep_results}(a), accumulating verified positive segments nearly doubles task completion from 20/60 without memory to 38/60, and adding structured failure guards lifts completion further to 48/60 by blocking repeated misaligned contacts. Combining both with direct verified dispatch, WB-RSI reaches 51/60 while cutting online execution time by over 60\% and reducing both agent queries and token usage.

\paragraph{Checked-program reuse.}
We further isolate checked skill-program execution against stepwise agent control on 24 unseen paired layouts sharing frozen round-10 \texttt{S+F} memories. Both arms solve all 24/24 tasks, with every invoked skill program passing internal postcondition and goal checks (Table~\ref{tab:rsi_multistep_results}(b)). Executing checked programs reduces agent calls, token usage, and online execution time across the board, halving manipulation-phase queries on multi-step box transfer.

\subsection{World-Model Planning}\label{wap-budget-validation}

To isolate online predictive planning without cross-episode memory, we evaluate 20 paired full-task box-transfer layouts featuring initially unseen targets, floor barriers, overhead clearance constraints, and timed lateral impulses. Both configurations use GPT-6.1 Sol over identical sensing, viewpoint, approach, grasp, transport, and in-scene recovery primitives; whereas the ablation queries the agent at every manipulation decision, WB-RSI instantiates \(p_\theta\) as an action-conditioned world model fine-tuned from Wan2.1-T2V-1.3B~\cite{wan2025wan} on 100 offline trajectories, scoring candidate grasp and transport rollouts locally via Eq.~\eqref{eq:wm-selection} and querying the agent only when forecasts fall outside \(\mathcal U_t^{\mathrm{supp}}\).

\begin{table}[t]
\centering\small
\setlength{\tabcolsep}{2pt}
\caption{Full-task world-model planning ablation on 20 paired box-transfer layouts with obstacles and external impulses (episodic memory disabled).}
\label{tab:wap_budget}
\begin{tabular*}{\columnwidth}{@{}l@{\extracolsep{\fill}}rrrr@{}}
\toprule
Configuration & Success $\uparrow$ & Calls $\downarrow$ & Tok. (k) $\downarrow$ & Online (s) $\downarrow$ \\
\midrule
w/o WM planning & 17/20 & 5.25 & 127.67 & 137.61 \\
WB-RSI (ours) & \textbf{18/20} & \textbf{4.15} & \textbf{58.16} & \textbf{119.63} \\
\bottomrule
\end{tabular*}
\end{table}

As shown in Table~\ref{tab:wap_budget}, WB-RSI improves full-task completion to 18/20 versus 17/20 without world-model planning while slashing LLM token consumption by 54.4\%, more than a $2.2\times$ reduction, alongside fewer agent queries and lower end-to-end online time inclusive of rollout inference. World-model planning thus offloads over half of the foundation-model token budget onto fast local evaluation while strengthening reliability under physical disturbances.

\subsection{Active Perception}\label{active-perception-validation}

We next test whether active perception reliably localizes and verifies manipulation targets that start entirely outside the egocentric camera view. On a fixed-base G1, torso-mounted RGB-D views are controlled via waist pitch and yaw; success requires an independently verified RGB-D measurement rather than pixel visibility alone.

\begin{figure}[t]
\centering
\makebox[0.49\columnwidth][c]{\footnotesize Without active perception}\hfill
\makebox[0.49\columnwidth][c]{\footnotesize With active perception}\\[2pt]
\includegraphics[width=\columnwidth]{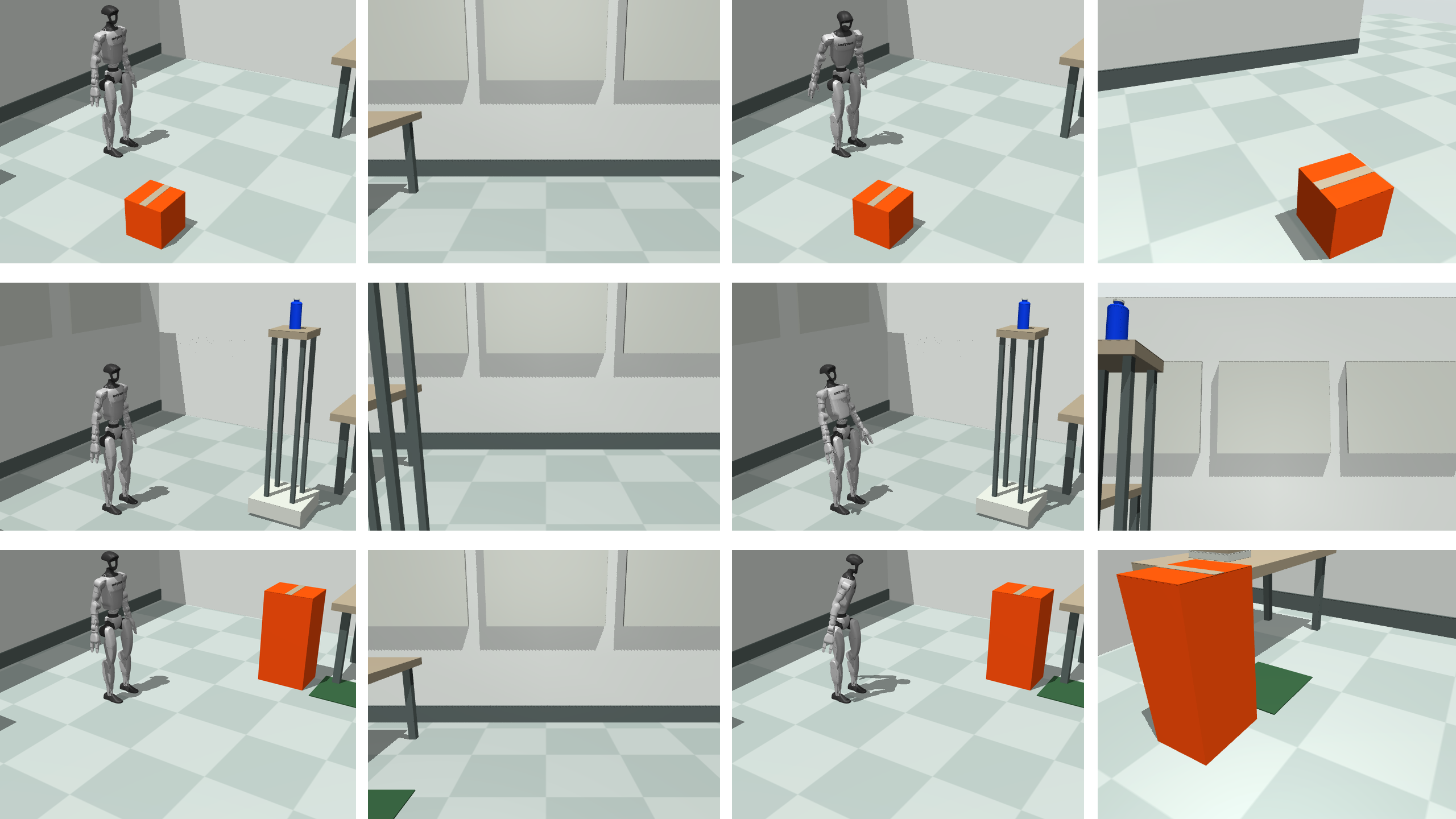}
\caption{\textbf{Active perception in MuJoCo.} Rows from top to bottom show a floor-level box, an elevated bottle, and a large box before pushing. Each condition shows an external view followed by the robot's camera view. Waist pitch and yaw bring initially unseen targets into view.}
\label{fig:active_perception}
\end{figure}

\begin{figure*}[t]
\centering
\includegraphics[width=\textwidth]{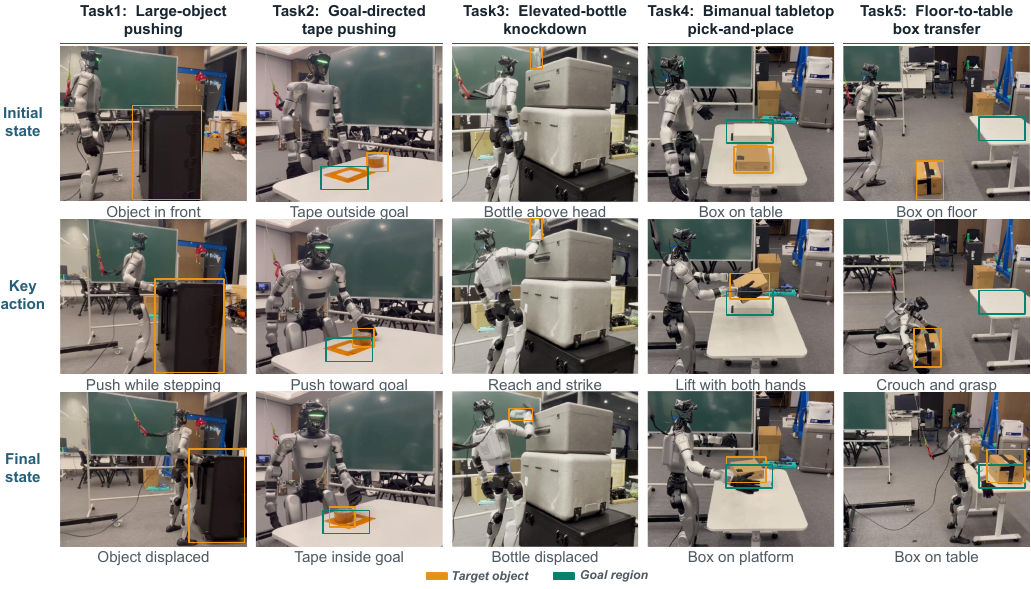}
\caption{\textbf{Real-robot execution examples.} Columns from left to right show large-object pushing, goal-directed tape pushing, elevated-bottle knockdown, bimanual tabletop pick-and-place, and floor-to-table box transfer (T1--T5). Rows show the initial state, key action, and final state. Orange boxes mark target objects; teal boxes mark goal regions.}
\label{fig:realworld}
\end{figure*}

\begin{table}[t]
\centering\small
\setlength{\tabcolsep}{2.5pt}
\caption{Active perception (AP) evaluation: (a) ablation across agent backbones (20 paired layouts) and (b) comparison against direct VLM viewpoint exploration (30 layouts; WB-RSI uses GPT 6 Astra).}
\label{tab:active_perception}
\smallskip\textbf{(a) AP ablation across Agent models (20 paired layouts)}\par\smallskip
\begin{tabularx}{\columnwidth}{@{}Yrrrr@{}}
\toprule
\textbf{Agent model} & w/o AP & With AP & Box & Bottle \\
\midrule
GPT 6.1 Sol & \textbf{0/20} & \textbf{15/20} & \textbf{6/10} & \textbf{9/10} \\
Kimi K3.1 Pro & \textbf{0/20} & \textbf{15/20} & \textbf{6/10} & \textbf{9/10} \\
DeepSeek V4 Pro & \textbf{0/20} & 14/20 & 5/10 & \textbf{9/10} \\
GPT 6 Astra & \textbf{0/20} & \textbf{15/20} & \textbf{6/10} & \textbf{9/10} \\
\bottomrule
\end{tabularx}
\par\medskip\textbf{(b) Agentic exploration (30 layouts)}\par\smallskip
{\footnotesize
\begin{tabularx}{\columnwidth}{@{}Yrrrr@{}}
\toprule
\textbf{Method} & Verified $\uparrow$ & Discovery $\uparrow$ & Calls $\downarrow$ & $T_{120}$ (s) $\downarrow$ \\
\midrule
GPT 6.1 Sol & 17/30 & 25/30 & 2.00 & 81.79 \\
Kimi K3.1 Pro & 26/30 & 29/30 & 2.37 & 62.81 \\
GPT 6 Astra & 26/30 & \textbf{30/30} & 2.93 & 46.63 \\
DeepSeek V4 Pro & 21/30 & 23/30 & 2.43 & 96.38 \\
WB-RSI & \textbf{27/30} & \textbf{30/30} & \textbf{1.30} & \textbf{25.04} \\
\bottomrule
\end{tabularx}
}
\end{table}

\paragraph{Ablation and exploration efficiency.}
Across 20 paired floor-box and elevated-bottle layouts in Table~\ref{tab:active_perception}(a), enabling active perception verifies 14--15/20 targets across four distinct foundation models, compared with 0/20 without viewpoint control. On 30 extended layouts that add large pushing cartons, we compare WB-RSI's structured six-view sweep and local RGB centering against four VLM baselines that command continuous waist angles directly under identical motion and verification budgets (Table~\ref{tab:active_perception}(b)). WB-RSI achieves the highest verification rate at 27/30 with complete 30/30 target discovery; relative to the matched GPT-6 Astra backbone, observation batching and local centering cut agent calls by 55.7\% and nearly halve failure-penalized acquisition time $T_{120}$.

\subsection{Generalization}\label{zero-shot-transfer}

We evaluate out-of-distribution generalization on ten paired box-transfer layouts where object poses, box dimensions, and destination heights shift beyond the training range of the world model and frozen source memory, compounded by floor barriers, overhead constraints, and timed lateral impulses. ReAct~\cite{yao2022react}, Reflexion~\cite{shinn2023reflexion}, ExpeL~\cite{zhao2024expel}, and default WB-RSI share GPT-6.1 Sol, identical RGB-D perception, and matched action and query budgets, with all resources kept frozen at test time.

\begin{table}[t]
\centering
\footnotesize
\setlength{\tabcolsep}{2pt}
\caption{Out-of-distribution box-transfer generalization on ten paired layouts with frozen resources (DeepSeek tokens reflect Qoder-reported usage).}
\label{tab:zero_shot_transfer}
\begin{tabular*}{\columnwidth}{@{}l@{\extracolsep{\fill}}rrrr@{}}
\toprule
Method & Success $\uparrow$ & Calls $\downarrow$ & Tok. (k) $\downarrow$ & Online (s) $\downarrow$ \\
\midrule
ReAct~\cite{yao2022react} & 7/10 & 4.60 & 110.70 & 68.74 \\
Reflexion~\cite{shinn2023reflexion} & 7/10 & 4.50 & 109.13 & 68.11 \\
ExpeL~\cite{zhao2024expel} & 6/10 & 4.60 & 111.75 & 68.59 \\
\midrule
WB-RSI (GPT 6.1 Sol) & \textbf{9/10} & 3.90 & \textbf{95.57} & 89.16 \\
WB-RSI (Kimi K3.1 Pro) & \textbf{9/10} & \textbf{3.70} & 185.27 & 91.51 \\
WB-RSI (DeepSeek V4 Pro) & 8/10 & 3.80 & 230.27 & 124.10 \\
\midrule
w/o WM planning & 8/10 & 4.30 & 103.95 & 64.99 \\
\bottomrule
\end{tabular*}
\end{table}

In Table~\ref{tab:zero_shot_transfer}, WB-RSI solves 9/10 out-of-distribution layouts with the matched GPT-6.1 Sol backbone, clearly outperforming ReAct and Reflexion at 7/10 and ExpeL at 6/10, while using fewer agent queries and fewer tokens. Replacing the closed-source agent with open-weight backbones maintains strong performance, reaching 9/10 with Kimi K3.1 Pro at the lowest query count and 8/10 with DeepSeek V4 Pro. Ablating world-model planning highlights a clear compute trade-off: active perception and source-memory orchestration alone already surpass prior agentic baselines at 8/10 with lower wall-clock latency, while enabling visual rollouts further trims agent queries and tokens and solves an additional constrained layout at the cost of video-prediction overhead.

\Needspace{140pt}

\subsection{Real-World Comparison}\label{realworld-qualitative}

Finally, we evaluate WB-RSI on a physical Unitree G1 across the five whole-body loco-manipulation tasks in Figure~\ref{fig:realworld}, comparing against three WAMs, three VLAs, and three humanoid agentic baselines over 20 trials per task, totaling 100 scored trials per method. All learnable methods share a common budget of 100 source trajectories per task mapped through validated G1 whole-body command adapters, and physical WB-RSI uses Kimi K3.1 Pro HighSpeed. The agentic baselines span LLM-synthesized whole-body control code in HuGo~\cite{choi2026hugo}, predicate-supervised planning in Cybo-Waiter~\cite{ren2026cybowaiter}, and VLM-mediated modular skills in Being-0~\cite{yuan2025being0}.

\begin{table}[t]
\centering
\footnotesize
\setlength{\tabcolsep}{2pt}
\caption{Physical Unitree G1 success rates over 20 trials per task (100 trials per method). Mean averages T1--T5.}
\label{tab:realworld_baselines}
\begin{tabular*}{\columnwidth}{@{}l@{\extracolsep{\fill}}rrrrrr@{}}
\toprule
\textbf{Method} & \textbf{T1} $\uparrow$ & \textbf{T2} $\uparrow$ & \textbf{T3} $\uparrow$ & \textbf{T4} $\uparrow$ & \textbf{T5} $\uparrow$ & \shortstack{\textbf{Mean} $\uparrow$} \\
\midrule
\multicolumn{7}{@{}l}{\textit{WAMs}} \\
Fast-WAM~\cite{yuan2026fastwam} & 0/20 & 10/20 & 6/20 & 13/20 & 0/20 & 29\% \\
OpenWAM~\cite{wang2026openwam} & 12/20 & 18/20 & 14/20 & 17/20 & 13/20 & 74\% \\
WB-WAM~\cite{qin2026wbwam} & 11/20 & 18/20 & 16/20 & 17/20 & 12/20 & 74\% \\
\midrule
\multicolumn{7}{@{}l}{\textit{VLAs}} \\
$\pi_{0.5}$~\cite{physicalintelligence2025pi05} & 12/20 & 16/20 & 14/20 & 15/20 & 14/20 & 71\% \\
DM0.5~\cite{dexmal2026dm05} & 12/20 & 17/20 & 10/20 & 16/20 & 12/20 & 67\% \\
GR00T N1.7~\cite{nvidia2026gr00tn17} & 13/20 & 14/20 & 11/20 & 15/20 & 13/20 & 66\% \\
\midrule
\multicolumn{7}{@{}l}{\textit{Agentic}} \\
HuGo~\cite{choi2026hugo} & 15/20 & 18/20 & 11/20 & 17/20 & 13/20 & 74\% \\
Cybo-Waiter~\cite{ren2026cybowaiter} & 17/20 & \textbf{19/20} & 12/20 & \textbf{19/20} & 14/20 & 81\% \\
Being-0~\cite{yuan2025being0} & 16/20 & \textbf{19/20} & 11/20 & 18/20 & 14/20 & 78\% \\
\midrule
WB-RSI (ours) & \textbf{18/20} & \textbf{19/20} & \textbf{18/20} & \textbf{19/20} & \textbf{16/20} & \textbf{90\%} \\
\bottomrule
\end{tabular*}
\end{table}

As shown in Table~\ref{tab:realworld_baselines}, WB-RSI achieves 90\% mean success across all 100 physical trials, outperforming the strongest agentic baseline Cybo-Waiter by 9 percentage points, the strongest WAMs by 16 points, and the strongest VLA by 19 points, while ranking best or tied-best on every task. Its largest margins emerge on elevated-bottle knockdown T3, where WB-RSI reaches 18/20 versus 11--12/20 for prior agentic methods, and on floor-to-table box transfer T5 at 16/20. Both tasks exemplify why tightly coupled perception, prediction, and memory matter on humanoids: targets start at extreme vertical angles outside the nominal camera frustum and demand coordinated torso pitch, crouch-to-stand height control, and contact-sensitive wrist trajectories. End-to-end VLA and WAM policies struggle when initial visual coverage is incomplete, whereas prior agentic baselines lack perception-gated permits and outcome-filtered whole-body references, leading to missed overhead strikes on T3 and unstable bimanual lifts on T5 that WB-RSI reliably avoids.

\Needspace{170pt}
\section{Conclusion}\label{conclusion}

We presented \textbf{WB-RSI}, an agentic self-evolving framework that enables humanoid robots to improve whole-body loco-manipulation directly from deployment experience. By coordinating an evolving experience memory and checked skill library with world-model planning and body-driven active perception over a unified whole-body command interface, WB-RSI resolves out-of-view targets prior to contact, evaluates candidate actions locally before execution, and progressively amortizes expensive foundation-model reasoning into fast, verified reuse. In simulation, harness evolution consistently improves multi-round task completion, query efficiency, and out-of-distribution generalization while keeping all foundation-model, world-model, and controller weights frozen. On a physical Unitree G1, WB-RSI achieves 90\% mean success across five whole-body loco-manipulation tasks, outperforming leading VLA, WAM, and agentic baselines. These results point toward deployment-driven harness evolution, grounded in active perception and predictive planning, as a practical path toward continually improving embodied intelligence.

\label{page:scientific_end}
\clearpage
\twocolumn[{
\section*{LLM Usage Statement}
\label{sec:llm_usage}
OpenAI ChatGPT/GPT and Codex supported manuscript preparation, implementation, experiments, and figures. The evaluated Agent collected simulation trajectories; outcomes were determined by physical measurements. The authors take full responsibility for all submitted content.

\par\bigskip
}]
{\small
\bibliographystyle{ieeenat_fullname}
\bibliography{references}
}
\clearpage
\maketitlesupplementary
\appendix
\setcounter{equation}{0}
\setcounter{table}{0}
\setcounter{figure}{0}
\renewcommand{\theequation}{S\arabic{equation}}
\renewcommand{\thetable}{S\arabic{table}}
\renewcommand{\thefigure}{S\arabic{figure}}
\renewcommand{\theHequation}{S\arabic{equation}}
\renewcommand{\theHtable}{S\arabic{table}}
\renewcommand{\theHfigure}{S\arabic{figure}}
\section{Implementation and Evaluation Conventions}
\label{supplemental-details}

This supplement follows the current main paper and its six experimental tables. Superseded experiments and presentation notes are collected in \path{oldversions.md}; their results are not pooled with the studies below. Unless stated otherwise, costs are per-attempt means including failures. Tokens count input plus output, with cached input included once; \textemdash\ denotes unavailable complete usage. Online time includes Agent response, execution, and world-model (WM) inference/transport. $T_B$ assigns $B$ seconds to failures and measured completion time to successes. Physical time, simulated time, and online wall time are distinct.

\subsection{Orchestration and command interface}
\label{supp:method-interface}\label{supp:method-supervisor}
The command adapter checks coordinate frames, joint and command limits, timestamps, and execution validity before passing accepted references to the frozen whole-body controller. Commanded references and achieved states are logged separately; stable holding commands persist during high-level inference. A candidate is rejected when its originating state is stale. Table~\ref{tab:states} summarizes execution priority and transitions.

\begin{table}[t]
\centering
\small
\setlength{\tabcolsep}{4pt}
\renewcommand{\arraystretch}{1.15}
\caption{Orchestrator modes and transition rules.}
\label{tab:states}
\begin{tabular}{>{\raggedright\arraybackslash}p{\dimexpr 0.22\columnwidth-2\tabcolsep\relax}>{\raggedright\arraybackslash}p{\dimexpr 0.32\columnwidth-2\tabcolsep\relax}>{\raggedright\arraybackslash}p{\dimexpr 0.46\columnwidth-2\tabcolsep\relax}}
\toprule
\textbf{Mode} & \textbf{Execution role} & \textbf{Transition} \\
\midrule
ASSESS & Update measurements, retrieve references, and evaluate checks. & Apply STOP, RECOVER, OBSERVE, REUSE, then PLAN priority. \\
REUSE & Execute an applicable reference, program, or neural skill. & Return after the local segment or enter RECOVER on an anomaly. \\
PLAN & Forecast finite candidates and check task-specific support and scores. & Select locally when supported; otherwise query the Agent. Verify execution from measurements. \\
OBSERVE & Acquire current task-relevant evidence through bounded observation actions. & Reassess after verification or exhaustion. Permit task execution only with valid current evidence. \\
RECOVER & Reestablish a locally valid execution state. & Resume after independent verification or stop at the limit. \\
STOP & Complete or terminate the episode. & Record success, timeout, intervention, or the stopping event. \\
\bottomrule
\end{tabular}
\end{table}

References are segmented at contact transitions, task-phase changes, and verified completion events. Retrieval uses goal, visual, body-state, and phase context, followed by applicability checks. An object-relative wrist trajectory is adapted by
\begin{equation}
T^{E,\mathrm{new}}_W=T^{O,\mathrm{new}}_W
\left(T^{O,\mathrm{ref}}_W\right)^{-1}T^{E,\mathrm{ref}}_W,
\label{eq:supp:method-wrist-adaptation}
\end{equation}
where $T_W^O$ and $T_W^E$ denote world-frame object and wrist poses. The adapted wrists are combined with body and navigation references. The same issued-command representation conditions offline prediction and deployment.

PLAN scores finite candidates from shared primitives. Predicted lift supports grasp ranking; predicted goal progress supports transport ranking. Unsupported rankings return selection to the Agent. Measurements verify every executed action. Execution deviation and stagnation have separate dwell times; hysteresis and mode-specific budgets limit repeated transitions. Observation and recovery checks precede resumption.

\subsection{Verified observations, outcomes, and skill versions}
\label{supp:method-observation}\label{supp:method-correction}\label{supp:method-skills}
A valid current observation permits the next task step. Further motion invalidates a step-specific permit; unresolved evidence blocks execution. Recovery restores a measured local condition without replacing the complete task goal, and resumes only after verification. Search, retries, and recovery consume the study's stated budgets.

Each segment keeps its starting context, goal, observation history, issued commands, termination event, and source. A failed grasp does not invalidate a successful approach; later placement does not relabel the failed grasp. Failure records form a subset of reference memory,
\begin{equation}
\mathcal B_{\mathrm{fail}}=\{(g_i,c_i,U_i,e_i,z_i,r_i,p_i)\}_i\subseteq\mathcal R,
\label{eq:supp:method-failure-memory}
\end{equation}
where $e_i$ is the observed event, $z_i$ its measured evidence, $r_i$ the recovery history, and $p_i$ the source record. Hypothesized causes remain distinct from observations. Verified recovery successes carry positive labels and join ordinary successes in $\mathcal D_{\mathrm{skill}}=\mathcal D_{\mathrm{succ}}\cup\mathcal D_{\mathrm{corr}}$; they are not a third outcome class.

A skill combines initiation conditions, a checked program, an executor, completion checks, and failure checks. Programs use object-relative parameters, a fixed tool interface, bounded retries, and measured postconditions. In the evaluated simulations, generated control programs are checked before evaluation; scored attempts select and parameterize them without regenerating code or training a neural executor. A neural executor can be fitted when enabled, using verified positive data and replay. Memory and skill versions update only between rounds. Regression checks accept a version before deployment, and provenance links each update to source evidence. The Agent, world model, and low-level controller remain frozen.

\section{Randomized Closed-Loop Deployment}
\label{supp:closed_loop}

\paragraph{Design and chronology.}
Main Table~1 evaluates three independent resource streams, each with ten box-transfer deployments per configuration. Each configuration starts with an empty episodic index and retains only its own measured outcomes. Resource versions at rounds 0, 5, and 10 are tested on ten locked layouts per stream without updating memory. The planned schedule contains 120 deployments and 360 evaluations. The final comparison uses 30 tests per configuration; repeated checkpoints on the same layouts are dependent learning traces.

\paragraph{Randomization and shared execution.}
Each stream has three initially visible, four search, and three search-plus-reference-shift test layouts. Reference X varies by $\pm3$\,cm around recurring workspace anchors. Target X adds $\pm2.5$\,cm in visible layouts or $\pm4$\,cm in search layouts. Shifted layouts have a persistent workspace bias of 8--11.5\,cm and fresh $\pm2$\,cm perturbations. Reference Y varies by $\pm1.5$\,cm and target Y by a further $\pm4$\,cm. Cube width, mass, and friction span $[.14,.19]$\,m, $[.24,.75]$\,kg, and $[.32,1.0]$. Destination X/Y/height span $[.55,.65]$\,m, $[-.04,.04]$\,m, and $[.36,.48]$\,m. Search starts use signed camera yaw magnitudes $[.85,1.25]$\,rad and pitch $[-.15,.15]$\,rad. Gripper home positions vary by $\pm2.5$\,cm horizontally and $\pm2$\,cm vertically. Candidate labels are shuffled. Initial states, observations, candidate controls, and prediction seeds are paired, with no outcome-based layout selection.

All configurations share primitives and public workspace references. A reference is locally applicable when measured center alignment is within 20\% of observed cube width and no matching failure rejects it. Otherwise grasp candidates use $-6$, $0$, and $+6$\,cm corrections. WB-RSI tries verified reuse, then frozen prediction, and queries the Agent for unsupported predictions or failed execution. Removing WM planning substitutes RGB geometry within this same routing; it does not force an extra query. Removing AP blocks view changes. Frozen memory retains the initial empty index, shared reference library, and current-episode feedback.

\paragraph{Sensing, prediction, and verification.}
Camera motion locates the target. A current RGB-D permit gates manipulation. Visible top/front surfaces estimate cube width and center; the destination is public calibration. Hidden spawn geometry, mass, friction, future contacts, and images are withheld. After acquisition, the camera returns to the world-model calibration view and records 61 causal RGB/control frames. The step-1900 predictor and twenty-step sampler forecast twenty frames over two seconds per candidate. The lift support/ranking rule in Section~\ref{supp:wap-multistep-current} is unchanged. No model fitting occurs.

The box remains free under contact dynamics. A grasp requires at least .20\,m measured rise and bilateral pad contact during the final .1\,s. Failed grasps trigger opening and retraction without resetting the object or velocity. Transfer, release, and measured goal verification follow a valid grasp. This contact proxy excludes G1 balance and SONIC locomotion.

\paragraph{Resources and budgets.}
Deployment records store context, action, local verification, source episode, and round. View retrieval matches workspace and measured initial-yaw sector; grasp retrieval matches workspace and image-relative reference displacement normalized by cube extent. Up to three neighbors are retrieved with feature tolerances .1 for views and .25 for grasps. Matching negative evidence blocks repeated-action reuse. Positive references still require fresh observation and contact checks. This is nonparametric reuse of fixed primitives.

Every episode permits one Agent query, three nonzero camera moves, two grasps, at most six neural candidate forecasts, 60 physics seconds, and a 180\,s success deadline. The policy uses at most one three-candidate batch before the first grasp. Agent calls are actual CLI launches. Online time includes sensing, control, forecasting, reasoning, and transport; scene allocation, shared warmup, and final archiving are excluded. Deployment acquisition costs are separate from evaluation.

\paragraph{Interruption accounting.}
An SSH outage interrupted one round-5 evaluation before an Agent process launched. Its record is retained; only the unexecuted schedule suffix continued from reconstructed resources. The affected layout is excluded for all four configurations at that checkpoint, leaving 29 tests/configuration; rounds 0 and 10 retain 30. No attempted episode is retried. Earlier development and an interrupted protocol-development version are excluded.

\begin{table*}[t]
\centering\small
\setlength{\tabcolsep}{4pt}
\caption{Checkpoint evaluation. Usage counts complete token reports; $\dagger$Frozen-memory tokens average 28 metered final tests, versus the full cohort elsewhere. WM counts neural forecasts. Round 5 has one paired transport exclusion.}
\label{tab:closed_loop_checkpoints}
\begin{tabular}{@{}rlrrrrrrr@{}}
\toprule
Round & Configuration & Success $\uparrow$ & Calls $\downarrow$ & Scans $\downarrow$ & Tok. (k) $\downarrow$ & WM $\downarrow$ & $T_{180}$ (s) $\downarrow$ & Usage $\uparrow$ \\
\midrule
0 & w/o AP & 9/30 & 0.00 & 0.00 & 0.00 & 0.00 & 128.54 & 30/30 \\
0 & w/o WM planning & 30/30 & 0.70 & 0.70 & \textemdash & 0.00 & 37.30 & 28/30 \\
0 & Frozen memory & 28/30 & 0.70 & 0.70 & \textemdash & 1.20 & 51.52 & 27/30 \\
0 & WB-RSI (ours) & 28/30 & 0.70 & 0.70 & \textemdash & 1.20 & 52.14 & 25/30 \\
\midrule
5 & w/o AP & 9/29 & 0.00 & 0.00 & 0.00 & 0.00 & 126.74 & 29/29 \\
5 & w/o WM planning & 29/29 & 0.17 & 0.69 & 3.07 & 0.00 & 16.39 & 29/29 \\
5 & Frozen memory & 27/29 & 0.69 & 0.69 & \textemdash & 1.24 & 51.88 & 26/29 \\
5 & WB-RSI (ours) & 29/29 & 0.21 & 0.69 & \textemdash & 0.31 & 20.51 & 27/29 \\
\midrule
10 & w/o AP & 9/30 & 0.00 & 0.00 & 0.00 & 0.00 & 128.50 & 30/30 \\
10 & w/o WM planning & 27/30 & 0.20 & 0.70 & 9.81 & 0.00 & 23.17 & 30/30 \\
10 & Frozen memory & 28/30 & 0.70 & 0.70 & $12.08^{\dagger}$ & 1.20 & 51.90 & 28/30 \\
10 & WB-RSI (ours) & 30/30 & 0.10 & 0.70 & 1.76 & 0.00 & 12.33 & 30/30 \\
\bottomrule
\end{tabular}
\end{table*}

\begin{figure}[t]
\centering
\includegraphics[width=\columnwidth]{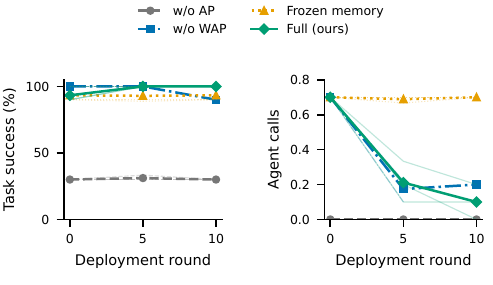}
\caption{Evaluation at deployment rounds 0, 5, and 10. Heavy lines are stream means; thin lines show available streams. The final w/o WM planning point uses its confirmed 27/30 aggregate; revised stream details are unavailable. The same layouts recur across checkpoints, with 29 tests/configuration at round 5 and 30 otherwise.}
\label{fig:closed_loop_learning}
\end{figure}

\begin{table*}[t]
\centering\small
\setlength{\tabcolsep}{6pt}
\caption{Final success by stream and difficulty. Each stream has ten tests; difficulty columns partition the same 30 tests (9 visible, 12 search, 9 shifted). w/o WM planning strata are unavailable.}
\label{tab:closed_loop_strata}
\begin{tabular}{@{}lrrrrrr@{}}
\toprule
Configuration & Stream 1 $\uparrow$ & Stream 2 $\uparrow$ & Stream 3 $\uparrow$ & Visible $\uparrow$ & Search $\uparrow$ & Shifted $\uparrow$ \\
\midrule
w/o AP & 3/10 & 3/10 & 3/10 & 9/9 & 0/12 & 0/9 \\
w/o WM planning & \textemdash & \textemdash & \textemdash & \textemdash & \textemdash & \textemdash \\
Frozen memory & 9/10 & 9/10 & 10/10 & 9/9 & 12/12 & 7/9 \\
WB-RSI (ours) & 10/10 & 10/10 & 10/10 & 9/9 & 12/12 & 9/9 \\
\bottomrule
\end{tabular}
\end{table*}

\begin{table}[t]
\centering\footnotesize
\setlength{\tabcolsep}{1.5pt}
\caption{Acquisition costs for 30 deployment episodes/configuration, separate from evaluation. Online time is measured wall time; $T_{180}$ penalizes failures.}
\label{tab:closed_loop_adaptation}
\begin{tabular*}{\columnwidth}{@{}l@{\extracolsep{\fill}}rrrrr@{}}
\toprule
Configuration & Success $\uparrow$ & Calls $\downarrow$ & WM $\downarrow$ & Online (s) $\downarrow$ & $T_{180}$ (s) $\downarrow$ \\
\midrule
w/o AP & 9/30 & 0.00 & 0.00 & 2.53 & 128.52 \\
w/o WM planning & 30/30 & 0.50 & 0.00 & 28.98 & 28.98 \\
Frozen memory & 28/30 & 0.70 & 1.10 & 40.62 & 49.59 \\
WB-RSI (ours) & 29/30 & 0.53 & 1.00 & 36.34 & 39.62 \\
\bottomrule
\end{tabular*}
\end{table}

\paragraph{Outcomes and interpretation.}
WB-RSI completes 30/30 final tests, including all 21 initially unseen targets, versus 28/30 with frozen memory and 9/30 without AP. The visibility mixture limits no-AP success to the initially visible 30\%; strata are reported separately. Current w/o WM planning results are 27/30, .20 calls, 9.81k tokens, and $T_{180}=23.17$\,s. WB-RSI reuses 18 view actions and 30 grasp corrections, with three Agent queries and zero final-test WM batches because deployment-time predictive selections have been consolidated into memory. Main Table~1 therefore compares acquired resources; main Table~3 separately tests online prediction.

During WB-RSI deployment, 30 neural candidate forecasts yield 29/30 task completions, 79 positive segments, and one failed grasp. That failure is followed by physical opening/retraction, but query exhaustion prevents successful recovery. Its negative record enters two later final-test decisions; this does not isolate the marginal effect of the failure guard. Both frozen-memory final failures exhaust the query budget before grasping.

\paragraph{Usage coverage and provenance.}
Final WB-RSI usage totals 52,736 tokens over all 30 tests. Frozen memory reports 338,374 tokens on 28/30 tests: 12.08k is a complete-case mean, while all other metrics retain 30 tests. Missing usage is not imputed. Revised w/o WM planning aggregate values are recorded in \path{data/current_publication/table1_author_update_20261117.json}; revised stream/difficulty breakdowns are unavailable and are not reconstructed from old exports. The raw inventory and source/model seals remain unchanged. Compact checkpoint exports are in \path{data/closed_loop/}; complete attempts, CLI receipts, resource snapshots, and continuation records remain in \path{G1-sim/result/closed_loop_v2/}.

\section{Memory Ablation and Checked-Program Reuse}
\label{supp:rsi-current}

\subsection{Ten-round memory mechanism comparison}
Main Table~2(a) compares no episodic memory, fixed initial memory, successful-record updates, success-plus-failure evidence, and WB-RSI over ten rounds of box transfer and bottle knockdown, with 60 matched tests per configuration. S and S+F use the same successful records, including verified recovery successes; S+F adds failed attempts as negative evidence. Sensing, AP, and primitives are shared. World-model planning is disabled. Separate source outcomes enter later rounds, and evaluation never updates memory.

\begin{table}[t]
\centering\footnotesize
\setlength{\tabcolsep}{2pt}
\caption{Mechanism comparison for main Table~2(a). Tokens are thousands per attempted task.}
\label{tab:supp:rsi_current}
\begin{tabular*}{\columnwidth}{@{}l@{\extracolsep{\fill}}rrrrr@{}}
\toprule
Configuration & Success & Calls & Scans & Tok. (k) & Online (s) \\
\midrule
No episodic memory & 20/60 & 4.17 & 3.00 & 97.13 & 70.20 \\
Fixed initial memory & 22/60 & 4.33 & 2.17 & 102.33 & 68.53 \\
S updates & 38/60 & 3.50 & 2.00 & 84.66 & 43.78 \\
S+F evidence & 48/60 & 3.50 & 2.00 & 84.95 & 34.27 \\
WB-RSI (ours) & 51/60 & 3.17 & 2.17 & 76.16 & 27.10 \\
\bottomrule
\end{tabular*}
\end{table}

Calls, scans, tokens, and online time average attempted test tasks, including failures. Online time is actual elapsed time. The old single-contact study's one-query, one-contact, and $T_{30}$ definitions do not apply. WB-RSI raises completion from 20/60 to 51/60, reducing calls by 24.0\% and test tokens by 21.6\%. Source acquisition is separate; no acquisition-inclusive saving is inferred from historical collection costs.

\paragraph{Record availability.}
These updated aggregates are author-confirmed publication values. The corresponding updated per-attempt records and full query/contact/time-budget specification are not included in the local export. \path{data/table2a_author_confirmation.json} records the pending raw evidence; its earlier no-memory scan value is superseded by the main PDF's 3.00. Current aggregates are in \path{data/current_publication/table2_a.csv}. Historical protocols are archived in \path{oldversions.md} and are not used to fill missing settings.

\subsection{Checked-program reuse on new layouts}
\label{supp:rsi-checked-reuse}
Main Table~2(b) uses 24 new paired layouts: twelve per task, with four per task from each of three source streams. Both arms share immutable round-10 S+F records, initial physical states/RGB, current feedback, and primitives. Registration classes $-18$, $0$, and $+18$\,cm and view sectors are balanced four per class per task. Target geometry, mass, friction, support height, goal, and camera jitter vary in recurring workspaces. Tests never update the source bank.

Each attempt permits eight queries, three contacts, six views, twelve primitives, 60 simulation seconds, 240 online seconds, and 60 seconds per request. Only availability of a checked reference program differs. The Agent must acquire a current observation and explicitly select the program. Box reuse verifies grasp before high transfer; bottle reuse has one contact step. A failed substep returns control, and recovery preserves the physical world. Programs are generated and checked before testing; no code generation or neural-executor fitting runs during scored attempts, and world-model planning is inactive.

\begin{table}[t]
\centering\footnotesize
\setlength{\tabcolsep}{1.5pt}
\caption{Checked-program reuse by task (twelve paired attempts/row).}
\label{tab:supp:rsi_checked_reuse}
\begin{tabular*}{\columnwidth}{@{}l@{\extracolsep{\fill}}rrrrr@{}}
\toprule
Configuration & Success $\uparrow$ & Calls $\downarrow$ & Scans $\downarrow$ & Tok. (k) $\downarrow$ & Online (s) $\downarrow$ \\
\midrule
\multicolumn{6}{@{}l}{Box transfer} \\
S+F evidence & \textbf{12/12} & 4.00 & \textbf{2.00} & 100.86 & 118.31 \\
WB-RSI (ours) & \textbf{12/12} & \textbf{3.00} & \textbf{2.00} & \textbf{74.19} & \textbf{86.76} \\
\midrule
\multicolumn{6}{@{}l}{Bottle knockdown} \\
S+F evidence & \textbf{12/12} & \textbf{3.08} & \textbf{2.08} & \textbf{73.38} & 92.04 \\
WB-RSI (ours) & \textbf{12/12} & 3.17 & 2.17 & 75.75 & \textbf{91.01} \\
\bottomrule
\end{tabular*}
\end{table}

Both arms complete 24/24 tasks with one contact each. All 24 selected programs execute successfully, with all 36 internal physical steps and final goals verified; there are no selected-but-not-started programs or partial failures. Box manipulation queries fall from 2.00 to 1.00; bottle manipulation uses one query in both arms, with one additional reuse view. Overall calls decrease from 85 to 74, tokens from 2,090,851 to 1,799,257, and measured online means from 105.17 to 88.88\,s. Simulation time is separate, and observed online timing includes provider/host variation.

All 159 scored requests have complete usage, with no request error or actor tool use. Four development smoke attempts are excluded. The shared historical source bank is unchanged and costs 272,249 tokens to collect; it is not recollected for this test. A scheduling interruption after fourteen completed attempts preserves every result; continuation admits only the remaining never-started cases, with no task or reply repeated. Full records are in \path{G1-sim/result/rsi_reuse_v1/formal_01/}; compact summaries, source seals, execution ledgers, and interruption audits are in \path{data/rsi_reuse_validation/}.

\section{Full-Task Frozen World-Model Ablation}
\label{supp:wap-multistep-current}

\paragraph{Paired tasks and controls.}
Main Table~3 uses twenty new paired box-transfer layouts (seed 2026111719): four open routes, eight floor barriers, and eight overhead constraints. Ten receive a predeclared lateral $\pm20$\,N impulse for .1\,s starting at 4.75--5.25 simulation seconds. Target pose, width, mass, friction, destination/support height, and initial camera vary. Three development worlds and a two-attempt integration smoke are excluded.

Both arms start with an unseen, unheld target and share RGB-D color/shape sensing, calibrated destination, initial state, fixed programs, and independent released/supported/settled placement checks. The manipulation overview becomes available only after actual acquisition. The Agent explicitly chooses approach before close/lift; there is no supplied successful-grasp prefix. Inspection, opening, retraction, and recovery preserve the same unreset physical scene. Episodic memory is disabled. Both arms use \texttt{gpt-6.1-sol/xhigh}, fresh contexts, current RGB/public measurements, and bounded within-episode history.

Each attempt permits eight queries, three grasps, six views, twelve primitives, 60 physics seconds, 240 online seconds, and 60 seconds per request. A response selects one primitive; the comparison does not allow unrestricted multi-action programs in one query. Started model errors/timeouts remain scored. Sensing, physics, native startup, queued responses, and prediction/transport are timed; scene allocation, global warmup, and final archiving are excluded. No new request, forecast, or physical action starts after the deadline. Bounded evidence collection for an in-flight reply cannot resume a timed-out task.

\paragraph{Predictor and local scores.}
The box predictor retains the step-1900 WanActionTextToVideo checkpoint and twenty-step sampler, trained from 100 trajectories split 80/10/10. Prediction receives only causal observed RGB, issued-control history, and proposed controls; candidate construction neither advances physics nor inspects future outcomes. WB-RSI permits one two-candidate batch per grasp/transport phase. Each selected action consumes the common execution budget. Neural forwards incur no LLM tokens but their inference/transport time is counted.

Grasp scoring uses generated orange-target centroids, calibrated camera rays, and the proposed constant-Y plane to estimate mean final-five-frame lift. The inherited support gate requires 18/20 detections per candidate, 4/5 final-frame detections, and .002\,m max--min score spread. These are relative estimates, not calibrated success probabilities. Transport scoring uses negative final-five-frame distance to the calibrated goal. A supported candidate needs 18/20 detections, 4/5 final-frame detections, and .15\,m signed predicted progress. One supported candidate may be selected; multiple candidates require a best-versus-second-best margin of .015\,m. Unsupported forecasts return selection to the same Agent interface. Physical guards and actual postconditions remain authoritative. The earlier single-selection experiments from which these definitions originate are in \path{oldversions.md}.

\begin{table*}[tbp]
\centering\small
\setlength{\tabcolsep}{5pt}
\caption{Full-task world-model planning ablation (twenty paired layouts/arm). Calls include search and manipulation. Primitive failures include unsuccessful views; WM time includes inference and transport.}
\label{tab:supp:wap_multistep_current}
\begin{tabularx}{\textwidth}{@{}Yrr@{}}
\toprule
Metric & w/o WM planning & WB-RSI \\
\midrule
Success & 17/20 & 18/20 \\
Open / floor / overhead success & 3/4 / 8/8 / 6/8 & 4/4 / 8/8 / 6/8 \\
Agent calls / task & 5.25 & 4.15 \\
Search queries / task & 1.95 & 1.95 \\
Approach/grasp/recovery-phase queries / task & 2.45 & 1.80 \\
Transport-phase queries / task & 0.85 & 0.40 \\
View actions / task & 1.95 & 1.95 \\
Grasp attempts / task & 1.15 & 1.20 \\
Opening/retraction actions / task & 0.05 & 0.15 \\
Selected primitive commands / task & 5.25 & 5.50 \\
Failed primitive checks / task & 1.15 & 1.20 \\
Blocked primitive commands / task & 0.00 & 0.00 \\
Input + output tokens / task (k) & 127.67 & 58.16 \\
Neural candidate forwards / task & 0.00 & 3.90 \\
Prediction/transport time / task (s) & 0.00 & 20.81 \\
Actual online wall / task (s) & 137.61 & 119.63 \\
Archived native-usage coverage & 20/20 & 19/20 \\
\bottomrule
\end{tabularx}
\end{table*}

\paragraph{Interruption and sensitivity.}
One infrastructure-invalid attempt stopped when a concurrent task terminated the resident predictor before a grasp request was written. Its prefix used two queries, 44.62k tokens, and 47.02\,s. Only that same frozen world restarted; task/model failures were not replaced. Thus forty scored attempts came from forty-one starts. Charging this prefix gives WB-RSI 4.25 calls and 121.98\,s per planned layout. Counting the interruption as failure without restart gives 17/20 for both arms. The interrupted worker's aborted-request count is unknown; completed replies and continuation-worker counters are audited separately.

\paragraph{Outcomes and evidence scope.}
WB-RSI completes 18/20 versus 17/20. Exact paired McNemar $p=1$ does not establish general success-rate dominance. Current author-confirmed all-twenty-attempt means are 4.15 calls, 58.16k tokens, and 119.63\,s, including 20.81\,s prediction/transport. Earlier native receipts cover nineteen WB-RSI attempts completely; this coverage is kept separate from the confirmed publication mean rather than changing its denominator. Source/checkpoint seals, per-world results, and native/physical/forecast ledgers are in \path{data/wap_multistep_v2/}; full evidence is in \path{G1-sim/result/wap_multistep_v2/formal_01/}. These are fixed-primitive contact-proxy tasks, not G1 balance or locomotion tests.

\section{Active Perception}

\subsection{Shared camera, motion, and verification}
\label{supp:active-perception}
The observation studies use MuJoCo 3.3.7 and the Unitree G1 with hands, a fixed pelvis, and forward-kinematic waist playback. The rigid torso camera has link-frame offset $(.11,0,.37)$\,m, $48^\circ$ vertical field of view, $12^\circ$ downward mounting, and $640\!\times\!480$ RGB-D output. There is no independent neck joint. Waist pitch/yaw changes the view while the base and other joints remain held. These studies stop at verified acquisition and exclude manipulation, SONIC locomotion, and dynamic balance.

Floor boxes use X/Y ranges $[1.00,1.15]$/$[-.98,-.72]$\,m and .30\,m cube sides. Bottle/support X/Y ranges are $[1.18,1.40]$/$[.65,.85]$\,m, with nominal support top 1.61\,m and height jitter $[-.04,.04]$\,m. Large pushing boxes use X/Y $[1.10,1.30]$/$[1.10,1.26]$\,m and half-sizes $[.21,.25]\times[.16,.20]\times[.43,.48]$\,m. Each scored target starts with zero visible pixels; paired conditions check sealed geometry fingerprints and initial RGB hashes without outcome-based replacement.

The shared local color/silhouette checker uses current RGB and back-projected rendered depth. A permit must match the episode and current step. Independent success additionally requires at least 100 true-target pixels, an eight-pixel image margin, and at least two of three sampled depth pixels on the true target. Renderer IDs are available only to initial-visibility audits and independent evaluation. Visibility alone is discovery; \texttt{finish} without valid evidence cannot succeed. Acquisition time is conditional on success, whereas all-attempt costs and failure-penalized $T_B$ retain failures.

The coarse search has yaw offsets $(0,.30,-.30,0,.45,-.45)$\,rad and pitch offsets $(.18,.18,.18,.32,.32,.32)$\,rad for floor boxes, with pitch signs reversed for bottles. Pushing uses yaw $(0,.30,-.30,0,.55,-.55)$ and the shallower pitch pair .16/.32\,rad. Interpolation is bounded by .5\,rad/s at 50\,Hz. One nonzero view change counts once even if split into transport chunks. Simulated reference duration is ticks/50 and is reported separately from online wall time.

\subsection{AP ablation across Agent models}
\label{supp:active-perception-ap20}
Main Table~4(a) uses twenty layouts, ten per profile. Ten retain the original diagnostic's geometry; ten are independently drawn with seed 2026111708. All four models execute both AP conditions on the same sealed worlds, giving 160 attempts, not eighty independently sampled worlds. AP order alternates within each model. Disabling AP blocks motion while retaining observation and verification. World-model planning and cross-episode updates are inactive.

Sol, Kimi, and DeepSeek request \texttt{gpt-6.1-sol/max}, \texttt{k3.1-pro-highspeed/max}, and \texttt{DeepSeek-V4-Pro/max}; the later Astra addition requests \texttt{gpt-6-astra/xhigh}. Each decision receives a fresh context containing current RGB, measured state, capability, budgets, and episode history, excluding hidden geometry and evaluator labels. Every attempt permits three calls, 600 reference ticks, 90\,s wall time, and at most 45\,s per call.

\begin{table}[tbp]
\centering\footnotesize
\setlength{\tabcolsep}{2pt}
\caption{AP ablation on twenty layouts/model. Costs include failures; acquisition medians condition on success.}
\label{tab:supp:active_perception_ap20}
\begin{tabular*}{\columnwidth}{@{}l@{\extracolsep{\fill}}lrrrr@{}}
\toprule
Agent model & AP & Verified $\uparrow$ & Calls $\downarrow$ & $T_{90}$ (s) $\downarrow$ & $T_{\mathrm{acq}}$ median (s) \\
\midrule
GPT 6.1 Sol & Off & 0/20 & 1.00 & 90.00 & \textemdash \\
 & On & 15/20 & 1.30 & 33.71 & 14.66 \\
Kimi & Off & 0/20 & 1.80 & 90.00 & \textemdash \\
 & On & 15/20 & 1.50 & 32.07 & 12.52 \\
DeepSeek & Off & 0/20 & 1.00 & 90.00 & \textemdash \\
 & On & 14/20 & 1.35 & 36.39 & 13.28 \\
GPT 6 Astra & Off & 0/20 & 1.00 & 90.00 & \textemdash \\
 & On & 15/20 & 1.30 & 43.37 & 26.91 \\
\bottomrule
\end{tabular*}
\end{table}

All failed responses and timeouts remain scored, including one DeepSeek no-answer response and a separate timeout. Qoder zero token placeholders mean unavailable usage. Timing includes client startup/network, rendering, and tool execution; immutable served-model build IDs are unavailable. The original 120 rows, manifest, costs, and no-answer audit are in \path{data/active_perception_ap20/}. The forty later Astra attempts, including one timeout, are in \path{data/table1_model_extension_v1/}. Raw traces are retained under \path{G1-sim/result/ap20_20261117/formal_a_connected_01/} and \path{G1-sim/result/table1_model_extension_v1/}.

\subsection{Direct exploration and refined view search}
\label{supp:active-perception-astra}
Main Table~4(b) uses thirty new paired layouts (seed 2026111710), ten per profile. Direct Astra, WB-RSI, and a fixed-scan diagnostic supply ninety attempts. Later direct Sol, Kimi, and DeepSeek substitutions add thirty each on the same worlds. The reported five Agentic methods therefore use 150 attempts; the separate fixed-scan diagnostic adds thirty. Neither world-model planning nor cross-episode memory runs.

Both Astra interfaces request \texttt{gpt-6-astra/high}; Sol/Kimi/DeepSeek request max effort. Each attempt permits six calls, 600 reference ticks, 120\,s after setup, and 45\,s per call. All share current RGB-D verification, motion limits, and initial-relative waist bounds of $\pm.55$\,rad yaw and $\pm.35$\,rad pitch. Direct exploration asks for numeric waist angles with actuator conventions and feedback, without named presets or a search macro. A fixed scan follows the same coarse bank with no Agent or image-based correction; it is a diagnostic of the preset program.

WB-RSI lets Astra invoke coarse search with up to three RGB-centering corrections after a visible but rejected view. For measured color-box center $(u,v)$ and $f=240/\tan(24^\circ)$,
\begin{align*}
\Delta\psi &= \operatorname{clip}[-.8\arctan((u-320)/f),-.15,.15],\\
\Delta\theta &= \operatorname{clip}[.8\arctan((v-240)/f),-.12,.12].
\end{align*}
Angles remain within the common waist bounds. At least 108 measured color pixels are required; repeated targets and corrections below .005\,rad are skipped. Every correction consumes motion/wall budget and receives a fresh check; thresholds are not relaxed and target truth is excluded.

\begin{table*}[tbp]
\centering\small
\setlength{\tabcolsep}{5pt}
\caption{Matched Astra exploration and fixed-scan diagnostic (30 attempts/condition). False permits count repeated checks; motion reference time is ticks/50.}
\label{tab:supp:active_perception_astra_current}
\begin{tabularx}{\textwidth}{@{}Yrrr@{}}
\toprule
Metric & \shortstack{Direct Astra\\exploration} & \shortstack{Fixed scan\\(no LLM)} & WB-RSI + Astra \\
\midrule
Verified acquisition $\uparrow$ & 26/30 & 22/30 & 27/30 \\
Floor-level box $\uparrow$ & 7/10 & 6/10 & 8/10 \\
Elevated bottle $\uparrow$ & 10/10 & 9/10 & 9/10 \\
Large box before pushing $\uparrow$ & 9/10 & 7/10 & 10/10 \\
Visible-target discovery $\uparrow$ & 30/30 & 30/30 & 30/30 \\
Agent calls / attempt $\downarrow$ & 2.93 & 0.00 & 1.30 \\
View changes / attempt $\downarrow$ & 2.53 & 5.07 & 4.10 \\
Motion reference time (s) $\downarrow$ & 2.24 & 4.20 & 1.80 \\
Online wall time (s) $\downarrow$ & 42.41 & 3.14 & 18.26 \\
$T_{120}$ (s) $\downarrow$ & 46.63 & 34.27 & 25.04 \\
Input + output tokens / attempt (k) $\downarrow$ & 66.68 & 0.00 & 29.32 \\
False-permit events (total) $\downarrow$ & 4 & 1 & 6 \\
Calls with usage $\uparrow$ & 88/88 & \textemdash & 39/39 \\
\bottomrule
\end{tabularx}
\end{table*}

\begin{table*}[tbp]
\centering\small
\setlength{\tabcolsep}{5pt}
\caption{Direct-exploration backbone substitutions on thirty shared layouts. Sol/Kimi/DeepSeek use max reasoning; WB-RSI uses Astra/high and repeats the reference results.}
\label{tab:supp:active_perception_direct_backbones}
\begin{tabularx}{\textwidth}{@{}Yrrrr@{}}
\toprule
Metric & \shortstack{Direct GPT 6.1 Sol\\exploration} & \shortstack{Direct Kimi\\exploration} & \shortstack{Direct DeepSeek\\exploration} & WB-RSI + Astra \\
\midrule
Verified acquisition $\uparrow$ & 17/30 & 26/30 & 21/30 & 27/30 \\
Floor-level box $\uparrow$ & 6/10 & 7/10 & \textemdash & 8/10 \\
Elevated bottle $\uparrow$ & 9/10 & 10/10 & \textemdash & 9/10 \\
Large box before pushing $\uparrow$ & 2/10 & 9/10 & \textemdash & 10/10 \\
Visible-target discovery $\uparrow$ & 25/30 & 29/30 & 23/30 & 30/30 \\
Agent calls / attempt $\downarrow$ & 2.00 & 2.37 & 2.43 & 1.30 \\
View changes / attempt $\downarrow$ & 1.47 & 2.07 & \textemdash & 4.10 \\
Motion reference time (s) $\downarrow$ & 1.55 & 1.68 & \textemdash & 1.80 \\
Online wall time (s) $\downarrow$ & 69.17 & 59.02 & \textemdash & 18.26 \\
$T_{120}$ (s) $\downarrow$ & 81.79 & 62.81 & 96.38 & 25.04 \\
Input + output tokens / attempt (k) $\downarrow$ & \textemdash & \textemdash & \textemdash & 29.32 \\
False-permit events (total) $\downarrow$ & 1 & 0 & \textemdash & 6 \\
Calls with usage $\uparrow$ & 53/60 & 70/71 & \textemdash & 39/39 \\
\bottomrule
\end{tabularx}
\end{table*}

WB-RSI verifies 27/30 versus Direct Astra's 26/30, with 55.7\% fewer calls and 56.0\% fewer tokens. Fixed scan is cheaper online but verifies 22/30. The Astra pair has 25 shared successes, two shared failures, two WB-RSI-only successes, and one direct-only success (exact McNemar $p=1$). This finite cohort supports reduced Agent interaction through batching and local feedback, without establishing general success-rate dominance or improved Agent weights.

\paragraph{Verification limits and interruptions.}
All three original conditions discover every target. On one bottle layout, local color bounds omit the nonblue cap: the local permit passes but the independent full-target margin fails. WB-RSI has six repeated false-permit events at that view; fixed scan has one. Direct Astra has four false permits on a different pushing layout. Independent failures remain failures, illustrating a limitation of color-region completeness.

The later Sol/Kimi arms contain 131 native calls, with 123 usage receipts and 17 timeouts; all model timeouts remain scored. One separate Sol attempt ended with worker SIGTERM before its third answer executed and is archived as infrastructure-invalid. A single audited restart on that layout yields the reported 17/30; counting the interruption as failure instead gives 16/30. The unchanged frozen protocol and all other outcomes are retained. Wall time includes provider waits and host load, so service comparisons do not isolate model inference latency. Missing usage is not zero.

The current DeepSeek result is the author-supplied update: 21/30 verified, 23/30 discovery, 2.43 calls, and $T_{120}=96.38$\,s. Unreported detailed metrics remain unavailable. Original Astra/fixed-scan records are in \path{data/active_perception_astra_refined/}; later model records are in \path{data/active_perception_direct_backbones/}, and current DeepSeek aggregates in \path{data/active_perception_deepseek_author/}. Full traces remain under \path{G1-sim/result/ap20_20261117/}. Earlier AP cohorts are archived separately in \path{oldversions.md}.

\section{RGB-D Box Generalization with Frozen Resources}
\label{supp:robust-box-generalization}

\paragraph{Cohort and scope.}
The primary comparison in Main Table~5 uses ten prospectively sampled box layouts with seed 2026111712, paired across ReAct, Reflexion, ExpeL, GPT-backed WB-RSI and w/o WM planning (50 attempts). Two later backbone substitutions add twenty attempts on the same worlds, as detailed below. All ten are retained without outcome filtering or replacement. The scope was restricted to boxes before box execution, retaining the unchanged box subset of the original sealed box/bottle plan. Its interrupted and completed bottle records remain archived and are not pooled here. A separately executed scripted feedback controller is excluded from the Agentic comparison by method role after execution; its ten records are also retained separately. This selection changes neither a layout nor an outcome, and does not establish superiority over non-Agentic controllers.

\paragraph{Scenes and collision physics.}
Target $x$ lies in $[-.490,-.435]\cup[.225,.280]$\,m, $|y|$ in $[.080,.095]$\,m, box width in $[.196,.214]$\,m, destination support height in $[.50,.55]$\,m and goal $x$ in $[.665,.690]$\,m. These extend beyond the original world-model collection ranges. Mass varies over $[.28,.68]$\,kg and friction over $[.40,.95]$. Four layouts have open routes, three a floor barrier with top height $[.73,.77]$\,m, and three an overhead barrier centered at $z=.98$\,m; barrier centers use $x=.43$, $y=.025$\,m. Five predesignated layouts receive an external $x$ force of $\pm25$\,N for 0.1\,s, starting at a sampled physics time in $[4.75,5.25]$\,s. The timing is independent of actions, method and outcome. Every modified scene is recompiled from XML before simulation, updating the static collision hierarchy. These are synthetic contact-proxy stressors, not a calibrated G1 disturbance or balance evaluation.

\paragraph{Observations and shared execution.}
A current RGB-D color-mask/shape fit uses calibrated cameras and an explicit orange upright-cube prior. No simulator target pose, target dimension, segmentation identity or hidden mass/friction enters actor inputs. Independent terminal checks may use simulator truth. All arms retain the same known goal, native primitives, automatic grasp-approach prefix, post-acquisition manipulation overview and physical feedback. These shared priors and motor routines remain substantial task structure; the comparison does not test end-to-end motor learning. Each primary-comparison Agent requests \texttt{gpt-6.1-sol/xhigh} through the same native CLI adapter with a fresh context and bounded episode history. ReAct has no episodic source memory; Reflexion and ExpeL use frozen source-language rules. WB-RSI and w/o WM planning share the same frozen source-view bank of ten other-task trajectories. No scored outcome updates any resource.

\paragraph{Prediction and budgets.}
The box world model remains at validation-selected step 1900 from the existing 100-trajectory dataset (80/10/10 split); no new fitting uses these worlds. Grasp candidates are ranked by predicted lift, transport candidates by predicted goal progress. Only causal RGB/control history and proposed commands enter forecasting. Unsupported rankings return selection to the Agent. At most one forecast batch per manipulation phase is allowed. Each attempt permits five actual Agent queries, seven primitive actions, 30\,s of physics, 600\,s online wall time and 60\,s per query. Dispatch order rotates within each layout, with at most three method arms concurrent and one neural worker. Wall time includes forecasting and transport, with startup logged separately; it should not be compared directly with older runs using a different concurrency policy.

\begin{table}[H]
\centering\footnotesize
\setlength{\tabcolsep}{1.5pt}
\caption{Additional metrics for main Table~5 (ten paired layouts/method). The revised GPT success count has no corresponding updated $T_{30}$. WM $\Sigma$ counts total neural forwards.}
\label{tab:supp:robust-box-generalization}
\begin{tabular*}{\columnwidth}{@{}l@{\extracolsep{\fill}}rrrrr@{}}
\toprule
Method & Success & Scans & Actions & $T_{30}$ (s) & WM $\Sigma$ \\
\midrule
ReAct & 7/10 & 2.50 & 4.60 & 16.817 & 0 \\
Reflexion & 7/10 & 2.50 & 4.50 & 16.724 & 0 \\
ExpeL & 6/10 & 2.50 & 4.60 & 18.585 & 0 \\
WB-RSI (GPT) & 9/10 & 2.40 & 4.90 & \textemdash & 40 \\
\midrule
WB-RSI (Kimi) & 9/10 & 2.40 & 4.40 & 12.929 & 38 \\
WB-RSI (DeepSeek) & 8/10 & 2.60 & 4.40 & 15.031 & 36 \\
\midrule
w/o WM planning & 8/10 & 2.30 & 4.30 & 14.565 & 0 \\
\bottomrule
\end{tabular*}
\end{table}

\paragraph{Results and latency trade-off.}
The current main table reports 9/10 for GPT-backed WB-RSI, versus 7/10, 7/10 and 6/10 for ReAct, Reflexion and ExpeL, and 8/10 for w/o WM planning. All 219 Agent queries in the primary native records have usage receipts. Tokens count input plus output once, including cached input only once. Forty WM forwards contribute 23.50\,s per WB-RSI attempt on average. Relative to w/o WM planning, calls and tokens decrease by 9.3\% and 8.1\%, while online wall time increases by 37.2\% due to video-forecasting overhead.

\paragraph{Later matched-layout backbone additions.}
We add twenty attempts on the unchanged ten box worlds, replacing only the Agent with Kimi K3.1 Pro Highspeed (\texttt{k3.1-pro-highspeed/max}) or Qoder CN 1.1.75 DeepSeek V4 Pro (\texttt{DeepSeek-V4-Pro/max}). The GPT cohort was already observed; these are later backbone substitutions rather than independent prospective confirmation. Two one-query native integration checks on the separate development layout are unscored. Initial observation hashes, source banks, action library, world-model checkpoint, sampler-seed rule and budgets remain matched. Two provider arms run concurrently within a world; one locked neural worker serializes complete forecast calls. Queueing is included in online time. Timing is descriptive across the earlier GPT run with a different concurrency policy. Kimi completes 9/10 and DeepSeek 8/10. The addition contains 2 failed native calls, of which 2 reach the common 60\,s limit; all remain task failures. An exclusive-create error in the progress logger halted scheduling after two completed attempts. Both native rows were preserved byte-for-byte; an explicit continuation executed only the remaining eighteen attempts without repeating a model request or task. The DeepSeek token cell uses the author-supplied Qoder calculation: 230.27k input-plus-output tokens per attempted task. Native zero placeholders are not used as token measurements. Complete receipts report 18.863613 Credits across 36/38 DeepSeek calls; missing or canceled usage is not imputed, and a full-cohort Credit mean is unavailable whenever any call is incomplete. Native trace audits and per-attempt exports are in \path{data/robust_box_backbones/}.

\paragraph{Token accounting.}
Main Table~5 reports input-plus-output tokens per attempted task, including failures. GPT-backed WB-RSI uses 39 queries and 955,740 tokens over ten tasks: 95.574k per task and 24.506k per query. DeepSeek uses 38 dispatches, or 3.80 calls per task, and the author-supplied Qoder calculation of 230.27k tokens per task. Credits are separate and are not converted into these token values. Cross-provider tokenization and accounting differ. Cached input is counted once; collection costs remain separate.

\paragraph{Provenance.}
The native run is \path{G1-sim/result/robust_generalization/box_formal_20261117_01}; the role-based presentation view is \path{G1-sim/result/robust_generalization/box_agentic_view_20261117_01}. Original records, sealed configuration and model hashes remain unchanged. The current publication uses the author-confirmed GPT success count and Qoder-calculated DeepSeek token value in main Table~5; these publication updates are recorded separately under \path{data/current_publication/}, alongside the archived numerical provenance. The superseded main-table text is consolidated in \path{oldversions.md}. Earlier bottle, instrumented-geometry, and cross-deployment backbone studies are archived in \path{oldversions.md}; none supplies additional rows for this comparison.

\section{Physical Comparison and Execution Interface}\label{supp:physical-interface}
The physical comparison reports one five-task evaluation on the G1, with 20 scored attempts per method--task pair. Mean is the unweighted average of the five task success rates; because each task has the same trial count, it also equals the pooled success rate over 100 attempts.

The comparison includes WAMs, VLAs, and the agentic humanoid methods HuGo, Cybo-Waiter, and Being-0. The VLA entries distinguish DM0.5 from $\pi_{0.5}$ and use the GR00T N1.7 release. The main comparison table cites the corresponding source methods and version-specific releases. The physical WB-RSI configuration uses Kimi K3 Pro HighSpeed as its Agent backbone with the WB-RSI harness and whole-body interface.

The five G1 tasks in the main paper's physical comparison table each have a budget of 100 source trajectories for task-specific adaptation, for 500 trajectories in total. A source trajectory is counted once even when its synchronized frames are converted into many training windows. The same task-specific source episodes are available to every learnable method. Any additional exploratory robot rollouts, generated data, or online updates are accounted for separately and cannot be treated as part of the common 100-trajectory budget. Scored physical attempts are excluded from model fitting, reference construction, and checkpoint selection.

Each source episode retains time-aligned camera observations, measured G1 state, issued actions, task instruction, and observed outcome. Model-specific preprocessing converts these records into the visual history, state representation, action horizon, and normalization expected by each released checkpoint. WB-RSI uses the recorded episodes both as outcome-labelled reference segments and as observed action--future-visual pairs for adapting the action-conditioned predictor; a changed command label alone does not supply the unobserved outcome of a counterfactual action.

The VLA and WAM baselines use the same source-episode pool under their own supervised post-training objectives. Source demonstrations and the primitive library are available for adapting the high-level agentic methods to the G1 interface; scored test outcomes are excluded from reference construction and adaptation.

At deployment, a model-specific adapter converts predicted actions or selected references to the G1 command representation supported by the whole-body interface. The interface checks coordinate frames, command and joint limits, timestamps, and execution validity before forwarding accepted blocks to the frozen controller. Issued commands and measured robot responses are logged separately; success is determined from measured task outcomes rather than model predictions. The 50 method/task rows are transcribed from the author-confirmed main Table~6 in \path{data/realworld_baselines.csv}. They record twenty trials per row and WB-RSI's 90/100 total. The table records the author-confirmed physical results; detailed per-trial traces and complete model-specific adapter/training configurations are not included in this source export.

\end{document}